# 3D Human Action Analysis and Recognition through GLAC descriptor on 2D Motion and Static Posture Images

## Mohammad Farhad Bulbul, Saiful Islam, Hazrat Ali*

*Dr. Md. Farhad Bulbul is with the Department of Mathematics, Jessore University of Science and Technology, Bangladesh (e-mail: farhad@just.edu.bd).*

*Saiful Islam is with the Department of Mathematics, Bangabandhu Sheikh Mujibur Rahman Science & Technology University, Bangladesh. (e-mail: saifulislambsmrstu@gmail.com).*

*Dr. Hazrat Ali is with the Department of Electrical and Computer Engineering, COMSATS University Islamabad, Abbottabad Campus, Pakistan (e-mail: hazratali@cuiatd.edu.pk ).*
*Corresponding author: Hazrat Ali, hazratali@cuiatd.edu.pk*

*Abstract*— **In this paper, we present an approach for identification of actions within depth action videos. First, we process the video to get motion history images (MHIs) and static history images (SHIs) corresponding to an action video based on the use of 3D Motion Trail Model (3DMTM). We then characterize the action video by extracting the Gradient Local Auto-Correlations (GLAC) features from the SHIs and the MHIs. The two sets of features i.e., GLAC features from MHIs and GLAC features from SHIs are concatenated to obtain a representation vector for action. Finally, we perform the classification on all the action samples by using the l2-regularized Collaborative Representation Classifier (l2-CRC) to recognize different human actions in an effective way. We perform evaluation of the proposed method on three action datasets, MSR-Action3D, DHA and UTD-MHAD. Through experimental results, we observe that the proposed method performs superior to other approaches.**

## I. INTRODUCTION

Research in human action recognition (HAR) is considered as one of the most interesting domains of computer vision. The action recognition system is being extensively applied in human security system, medical science, social awareness, and entertainment [1], [2], [3], [4].. Indeed, to develop an applicable action recognition system, researchers still need to win against the odds due to diversity in human body sizes, appearances, postures, motions, clothing, camera motions, viewing angles, and illumination. Besides, difficulty level of the action recognition task increases by the intra-class variations and inter-class similarities amongst actions. In the early stage, the human action recognition system was developed by researchers based on RGB data [5], [6], [7], [8]. But, the RGB data based recognition methods are less effective to address the aforementioned challenges [1], [9]. In addition, the RGB action video sequences can merely encode the 2D action data tempted with the lateral movement of the scene parallel to the 2D plane. In this situation, one requires to handle the deficiency of 3D information in traditional images. Besides, a significant number of hardware resources are required to deploy the action recognition system as a result of computationally rigorous image processing and computer vision algorithms. Consequently, recognition of actions accurately is considered as a challenging task.

Nevertheless, with the introduction of the depth data sensors, significant progress on action identification process has been observed during the last several years. The outcomes of a depth sensor are called depth maps. Depth maps capture distance between object's surface and the sensor's viewpoint [10]. The depth map pixels are actually standardized depths in the visual scene. Here, it is noted that depth maps are unaffected in lighting situations as well as texture changes compared to RGB images [11]. Depth images provide body, shape, structure information and 3D motions of the subjects in the scene. Moreover, the issues of human localization and segmentation are simple while processing depth images rather than the RGB images [12]. Examples of depth sequences for *forward punch, hammer* (the hammer action refers to the sport of throwing the hammer) and *side kick* actions are illustrated in Figure 1. Besides, the information of human skeleton is obtained by depth maps, which provide additional information in action labeling [13]. Roughly speaking, depth video data has some notable aspects making it more a preferred choice over the RGB data, such as action recognition regarding inferior lighting environments and even in darkness, approximating standardized depth in a scene, being stable to color and texture changes, and solving the silhouette issue in body posture [11]. These sensors also eliminate various difficulties in computer vision research, e.g., image backward scene removal and object segmentation.

By the above discussion, we are motivated to propose the depth data based action recognition system in this paper. Therefore, the motion and static information of an actor is collected in three motion history images (MHIs) and in three static history images (SHIs) corresponding to a depth action video. The above motion and static images are calculated by employing the 3D Motion Trail Model (3DMTM) on each depth map sequence. The 3DMTM basically constructs the MHIs and SHIs by taking the front, side and top projections of each depth map and accumulating consecutive differences of these projections along a specific view. The obtained static and motion posture images are passed to the GLAC [14] descriptor to encode the texture information for describing an action. The GLAC on the MHIs and SHIs generates six feature vectors and those vectors are fused with an optimistic strategy to get the final action representation vector. The gained action vector is fed to the l2-regularized Collaborative Representation Classifier (l2-CRC) to label a query action sample. The proposed method is graphically illustrated in Figure 2.

The main contributions of this paper are summarized as follows:

1.  The 3DMTM algorithm is employed on each depth map sequence to compute the MHIs and SHIs.
2.  The obtained MHIs and SHIs are fed into the GLAC descriptor and the output feature vectors are combined into a single action representation vector.
3.  The action feature vector is passed to the l2-CRC to classify the action sample.
4.  The proposed system is comprehensively assessed based on MSRAction3D [15], DHA [16], and UTD-MHAD [17] action datasets. We make comparison of the recognition outcome with hand-crafted feature based methods as well as deep learning methods. Overall experimental assessment indicates that the proposed approach achieves superiority over the aforementioned approaches (i.e., hand-crafted feature based methods and deep learning methods)

The rest of the paper is organized as follows. We consider the literature review in Section 2. The proposed system is discussed in Section 3 extensively. Section 4 presents the experimental evaluation of the method. Finally, we draw conclusion and outline future work in Section 5.

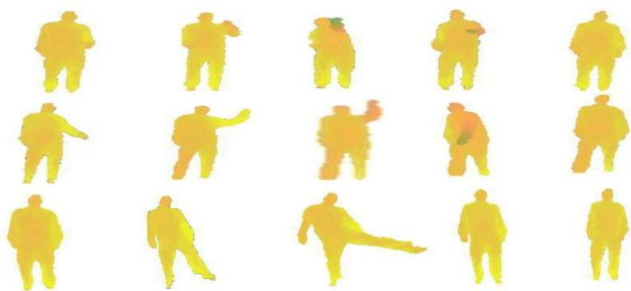

Figure 1. Example action sequences for *forward punch, hammer and side kick* actions

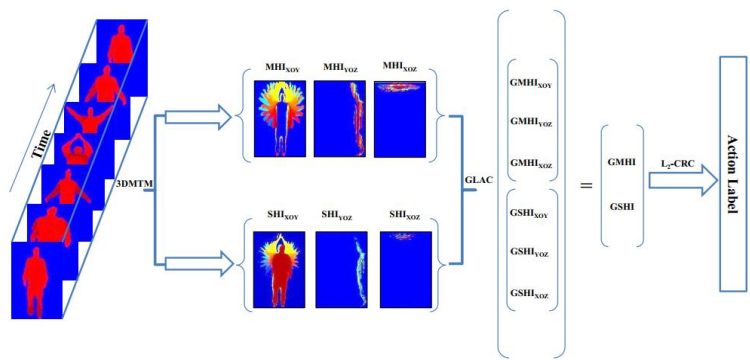

Figure 2. Framework of the proposed action recognition system

## II. Related Work

Researchers in computer vision have been greatly fascinated by the diversity in feature extraction and representation strategy in depth data based action recognition system. Thus, depth action sequences have been described through multiple action features encoding techniques such as 3D point cloud [15], projected depth images [18], spatio-temporal interest points [19], and skeleton joints [20]. In [15], three-dimensional interest points are gathered from all the depth maps of a depth action video to illustrate human action. The space-time occupancy patterns (STOP) are proposed by [21], [22] to represent human actions simply. Besides, Wang *et al.* introduced the random occupancy patterns (ROP) features in [23].

In [19], the space-time significant points (STIPs) in depth action sequences are captured after filtering noise. Similarly, the depth cuboid similarity feature (DCSF) was applied to designate the local 3D depth cuboid in action recognition [19]. After being motivated with the motion energy images (MEI) [20] of motion history images (MHIs), 2D projections (three projections) of each depth image were taken in 3D Euclidean space. Then the subtraction operations between consecutive projections were carried out to construct depth motion maps (DMMs) [18]. Another process, which is known as the histogram of oriented gradients (HOG) features [24] was then exploited from the DMMs as global representations of a depth action sequence. As a result, depth motion maps effectively transformed the problem as 3D to 2D mapping. Also, the strategy of gaining DMMs [25] was altered to bring the computational simplicity in the action recognition system so as to implement it in real-time situation. In [26], the local binary pattern (LBP) [27] algorithm was employed on the multiple overlapping chunks produced on DMMs to improve the classification power of the recognition system. Furthermore, HOG features in contourlet sub-bands (generated from DMMs) were extracted by [28]. In [29] and [30], DMMs-based texture and edge features were fused to increase the discriminatory power of obtained features. The 2D and 3D auto-correlation features were also captured from DMMs and were fused to distinguish depth actions [31], [32]. To enhance the DMMs, multi-temporal DMMs were computed and texture features were extracted by Chen and others in [33]. Also, 3D histograms of texture (3DHoTs) were used to capture dominant features from a depth action sequence for human action recognition [34]. The frontal, side, and top planes' projections in Cartesian plane were derived by 3DHoTs. Another classifier was proposed, which is referred to as multi-class boosting classifier (MBC), to robustly capture various types of action characteristics in an integrated action classification system. To capture the pixel-level shape motion cues about the complex joints, another descriptor was introduced by using a histogram of oriented 4D normal in [35]. This actually captured the distribution of the surface normal orientation in the 4D space of time, depth, and spatial coordinates. In [36], a novel system was proposed by merging salient depth maps (SDM) and binary shape map (BSM) feature vectors. For introducing a new method, locality-constrained linear coding (LLC) based action recognition algorithm was proposed in [37]. The method captured information of human actions in spatio-temporal subsequences of videos. The main experiment of that paper was completed with an $\ell_2$ regularization classifier as well as a linear SVM. In addition, an action categorization pipeline by using hierarchical 3D Kernel Descriptors from depth image sequences was described in [38]. In the approach, match kernel (EMK) was employed for classification for the next level of hierarchical structure.

Skeleton joints are gained from depth maps to represent a human body compactly. Based on those joints, several action classification methods have been developed by learning the correlation among action classes and 3D body-part joints from action depth maps. For example, the pairwise differences of 3D joint positions of a subject in a depth frame and the temporal differences about every depth map were calculated to illustrate human actions [39]. The classification results of the method in [39] were not promising since the extracted 3D skeletal joints were not capable to encode all of the dominant characteristics of an action. In that year, the above method was further improved by using histogram-based features [40]. These extracted features were taken around every joint in each depth map. Another framework for recognizing actions based on histograms of 3D joint locations (HOJ3D) was applied in [41]. This approach essentially encoded spatial occupancy information with respect to the skeleton root. Moreover, a genetic-based evolutionary algorithm was used to determine the optimum subgroup of skeleton joints [42]. It was also treated the topological construction of the skeleton to enhance the action classification outcomes. In fact, a binary vector was taken into account, where individual gene defined the attention or no attention of a specific action attribute. The filter and wrapper models were employed for its deployment. However, the computational complexity and the prompt convergence resulted in shortcomings of the method. Besides, the fitness measurement of a significant number of recognition outcomes was needed to attain the ultimate result through the wrapper-based evolutionary method. Actually, the association of the computation of a particular fitness through a learning and classification procedure took huge time for the entire evolution. Additionally, when the evolutionary results were collected in a local minimum and an expected outcome was not gained, then the early convergence occurred. A non-parametric moving pose (MP) approach for low-latency action identification was reported in [43]. For evaluating the method, a KNN algorithm was employed, which considered the temporal position of a specific frame inside the action sequence. A Fisher kernel (FK) exemplification was utilized to represent the skeletal quads as well as a Gaussian mixture model was learnt from learning data to encode the relative position of the joint quadruples of human skeleton, in [44]. Another joint representation and recognition model process by combining with multi-perspective and multi-modality categorized for 3D action recognition was described by [45]. In the work, the authors constructed a difference between motion historic images, then proposed multi-perspective projections for depth and color image sequence. The noise-robust actionlet ensemble model was presented in [46] to improve the action classification. The system considered the interaction between the subsets of human body joints. Another skeletal representation based on three-dimensional geometric linking among different body segments was proposed in [47].

For looking variations in skeleton joints related systems, 3D joint features were combined with spatio-temporal features [48]. The spatio-temporal features were captured on the color action video employing the center-symmetric motion local ternary pattern. Sometimes, the skeleton joint features are also incorporated into the depth image relevant features. As an example, Rahmani *et al*., incorporated the joint displacement histograms, joint movement occupancy volumes with the 4D depth, depth gradient histograms in [49].

In [1], the recognition was also improved by a blending pipeline with two different data sensors consisted of a depth sensor (a Kinect sensor) and a wearable inertial sensor (accelerometer). Individual attributes are captured from the action data obtained by the two sensors. The features of two different data corresponding to same action are fused from two different perspectives, i.e., feature based fusion and decision based fusion. In the method, the enhanced recognition accuracies were gained by utilizing action features from data of these two different modality devices mutually in comparison to the conditions when every device was utilized independently.

Besides the handcraft features based methods discussed above, all the deep learning models learn the action characterization from raw action data and properly compute the extreme level semantic action representation. In [50], Wang et al. introduced a deep model, where merely small-scale CNNs were required but exhibited superior performance with low computational complexity. In another method, DMM-Pyramid architecture was introduced for preprocessing the depth action videos [51]. In fact, the architecture is capable to preserve a part of temporal ordinal information from the depth sequence. In the system, the convolution operation was advocated to exploit spatial and temporal characteristics from raw action data spontaneously, and DMM to DMM-Pyramid was extended. Afterward, the raw depth action sequences were passed to convolutional neural networks with 2D and 3D architectures. In [62], by employing a video domain translation-scale invariant image mapping technique, 3D skeleton videos are mapped into skeleton color images and a multi-scale dilated convolutional neural network (CNN) adopts those images as inputs to classify them into distinct action categories. Multiple data augmentation approaches are considered to increase the generalization and robustness of this system. Lei *et al.* [63] characterized human actions through Spatio-Temporal Interest Points (STIP) features by following two-stage action modelling strategy. In the first stage, super-pixel Gaussian Mixture Model (GMM) is established to remove noise from the extracted STIP local features and individual class based codebook is constructed to obtain the basis for further inter-class feature collection. The spatio-temporal pyramid model (STPM) is then developed to describe spatial temporal correlation between those features. It should be noted the STPM yields a high dimensional features space. In the second stage, the combination of the Linear Discriminant Analysis (LDA) and the k-means clustering algorithm is employed in the STPM features space to gain high-level codebook and its corresponding feature representation.

Zhang *et al.* [64] introduced susceptibility weighted imaging to scan the subjects to detect the Cerebral Micro Bleed (CMB) voxels within brain. Then, to solve the accuracy paradox caused from the imbalanced data between CMB voxels and non-CMB voxels they used under sampling. As well as, they developed a seven-layer deep neural network (DNN). By combining both parametric rectified linear unit (PReLU) and dropout techniques. Zhang *et al.* [65] proposed an improved 10-layer convolutional neural network including 7 convolution layer and 3 fully connected layers. They collected 681 healthy control brain slices and 676 multiple sclerosis brain slices to complete their experiment.

In [66], Elmadany and others proposed a skeleton feature descriptor named Bag of Angles (BoA) and a depth feature descriptor for depth videos called Hierarchical Pyramid DMM Deep Convolutional Neural Network (HP-DMM-CNN). To enhance human actions Zeng *et al.* [67] proposed another process via structural average curves analysis from action samples.

The above discussion motivates this work on depth data oriented recognition system due to its superiority over other data based methods. Our work mainly focuses on the strategies of discriminative feature extraction and action representation with the extracted set of features. In feature extraction, this paper emphasizes on the motion as well as static image based feature extraction whereas previous methods extracted features from the motion images only. Overall, regarding the proposed system, our main hypothesis is that the inter-class similarity and the intra-class variation issues are considerably addressed when we encode the action features from the MHIs and SHIs and access them jointly to represent an action instead of using alone.

## III. PROPOSED RECOGNITION SYSTEM

In this section, we present our approach by a comprehensive discussion on feature extraction, action representation and classification techniques. Algorithm 1 describes our recognition system concisely. To elaborate the proposed method, the corresponding flowchart is shown in Figure 3.

---

***Algorithm 1: Proposed algorithm to address human action recognition issue***

| | |
|---|---|
| 1 | ***Input:*** *The input depth sequence is utilized to calculate MHIs ($MHI_{XOY}, MHI_{YOZ}, MHI_{XOZ}$) and SHIs ($SHI_{XOY}, SHI_{YOZ}, SHI_{XOZ}$) based on Eq. (1), (2) and (3).* |
| 2 | *The GLAC features are extracted from all the MHIs and SHIs via Eq. (5) and (6) to form **$GMHI$** = $[GMHI_{XOY}; GMHI_{YOZ}; GMHI_{XOZ}]$ and **$GSHI$** = $[GSHI_{XOY}; GSHI_{YOZ}; GSHI_{XOZ}]$.* |

| 3 | *The **GMHI** and the **GSHI** are combined to gain a single vector.* |
|---|---|
| 4 | *Pass the training feature set $P = \{p_i\}_{i=1}^n$, class label $c_i$ for class partition, test sample $S_t$, $\mu$, $C$ (number of action classes)* |
| 5 | *Calculate $\widehat{\beta}$ using Eq. (10)* |
| 6 | ***for all** j \in C$ **do*** |
| | *partition $P_j$, $\widehat{\beta_J}$* |
| | *Calculate $q_j = \left\| S - P_j\widehat{\beta_J} \right\|_2$* |
| | ***end for*** |
| | *Decide class ($S$) through Eq. (11)* |
| 7 | ***Output:** class ($S$)* |

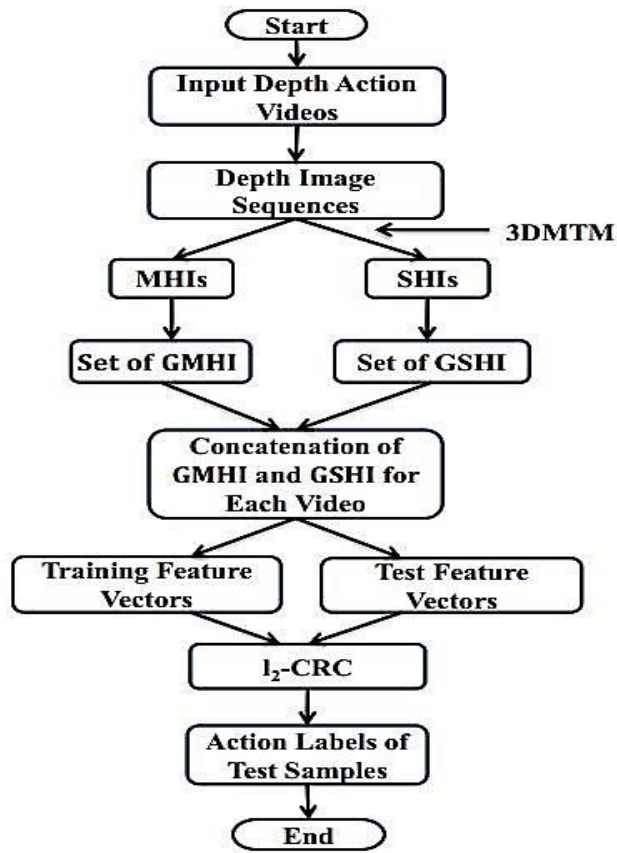

Figure 3. Flowchart of the proposed method

### A.  Features extraction

Our method extracts action features by employing the GLAC [14] on the MHIs and SHIs for each depth action video. Actually, the MHIs gather the motion information whereas the SHIs capture the motionless pose cues, monotonous movements and monotonous unmoving cues successfully [53]. It is clear that the MHIs and the SHIs store the complementary information of an action. However, the MHIs and SHIs are computed by passing an action video to the three-dimensional Motion Trail Model (3DMTM) [53]. In our problem, about each action video, the 3DMTM outputs the three MHIs as $\{MHI_{XOY}, MHI_{YOZ}, MHI_{XOZ}\}$ and the three SHIs as $\{SHI_{XOY}, SHI_{YOZ}, SHI_{XOZ}\}$ corresponding to three two-dimensional Euclidean faces. Figure 4 shows example of SHIs and MHIs computation corresponding to the *horizontal wave* action. The 3DMTM uses the motion update function $\varphi_M(x, y, t)$ and static posture update function $\varphi_S(x, y, t)$ to state the motion and motionless situations of an actor. Those functions are defined as:

$$\varphi_M(x,y,t) = \begin{cases} 1 & if P_t > \zeta_M \\ 0 & otherwise \end{cases}$$
$$\varphi_S(x,y,t) = \begin{cases} 1 & if d_t - P_t > \zeta_S \\ 0 & otherwise \end{cases} \qquad (1)$$

where $(x,y)$ is the position vector of a pixel at a time $t$. The $d_t = \{d_i\}_{i=1}^T$ denotes the sequence of depth images whereas $P_t = \{P_j\}_{j=1}^T$ indicates a sequence of differences between two successive depth frames. Also, the $\varphi_M(x,y,t)$ and $\varphi_S(x,y,t)$ require $\zeta_M$ and $\zeta_S$ threshold values to become concern about the motion and motionless portions within consecutive frames. So, the MHI $F_M(x,y,t)$ can be calculated using $\varphi_M(x,y,t)$ as

$$F_M(x,y,t) = \begin{cases} T & if \varphi_M(x,y,t) = 1 \\ F_M(x,y,t-1) - 1 & otherwise \end{cases} \qquad (2)$$

Following similar fashion, the SHI $F_S(x,y,t)$ is gained through $\varphi_S(x,y,t)$ by

$$F_S(x,y,t) = \begin{cases} T & if \varphi_S(x,y,t) = 1 \\ F_S(x,y,t-1) - 1 & otherwise \end{cases} \qquad (3)$$

It is worth to mention that the average motion history image (AMHI) and average static posture history image (ASHI) can also be generated by the 3DMTM but the AMHI and the ASHI decrease the recognition outcome [54]. As a result, we exclude them in our work. However, for intuitively stating the GLAC implementation on the MHI/SHI, let $I(x,y)$ stands for the MHI/SHI. At each pixel of $I(x,y)$, the gradient vector $\left(\frac{\partial I}{\partial x}, \frac{\partial I}{\partial y}\right)$ can be obtained by using the image gradient operators such as Roberts, Sobel, Kirsh, and one-dimensional derivatives ($[-1, 0, 1]$). The magnitude of gradient vector is expressed by $m = \sqrt{\left(\frac{\partial I}{\partial x}^2 + \frac{\partial I}{\partial y}^2\right)}$ and the orientation angle could be formulated as $\theta = \arctan(\frac{\partial I}{\partial x}, \frac{\partial I}{\partial y})$. The angle $\theta$ is coded by D orientation bins with the voting weights to the neighboring bins to form a D-dimensional gradient orientation (G-O) vector $\boldsymbol{g}$. The vector is basically in sparse form. By $\boldsymbol{g}$ and $m$, the $K^{th}$ order auto-correlation gradient function could be written as:

$$F(d_0, \dots d_K, \boldsymbol{a}_1 \dots \boldsymbol{a}_K) = \int w[(m(\boldsymbol{r} + \boldsymbol{b}_1), \dots, m(\boldsymbol{r} + \boldsymbol{b}_K)] g_{d_0}(\boldsymbol{r}) g_{d_1}(\boldsymbol{r} + \boldsymbol{b}_1) \dots g_{d_K}(\boldsymbol{r} + \boldsymbol{b}_K) d\boldsymbol{r} \qquad (4)$$

where $\boldsymbol{b}_1, \boldsymbol{b}_2, \dots, \boldsymbol{b}_K$ are called as shifting vectors with respect to the location vector $\boldsymbol{r} = (x,y)$ of a pixel in $I$, $g_d$ means the $d^{th}$ member of $g$ and $w(.)$ is a weighting function. Our experimentations employ $K \in \{0,1\}$, $a_{1xy} \in \{\pm \Delta r, 0\}$ and $w(.) \equiv \min(.)$ in Section 4 by [14].

In $K \epsilon \{0,1\}$, the GLAC features are written as follows:

$$\textbf{0}^{th} \textbf{ order GLAC: } \boldsymbol{F_0} = \sum_{r \in I} m(r) g_{d_0}(r) \qquad (5)$$

$$\textbf{1}^{st} \textbf{ order GLAC: } \boldsymbol{F_1} = \sum_{r \in I} \min[(m(r), m(r + \boldsymbol{b}_1)] g_{d_0}(r) g_{d_1}(r + \boldsymbol{b}_1)] \qquad (6)$$

Based on Equation (5) and Equation (6), the spatial auto-correlations, amongst local gradients over the gradient magnitude image (i.e., image of m) are calculated by using mask patterns as shown in Figure 5. There is a single mask pattern for Equation (5) and four independent patterns for Equation (6) while avoiding the duplicate patterns. For the $0^{th}$ order GLAC in Equation (5), summation is taken to only two non-zero elements of $g$ with weight $m$ about a pixel at $r$. For the $1^{st}$ order GLAC, summation of products in Equation (6) is considered to non-zero elements of $g(r)$ and $g(r + \boldsymbol{b}_1)$ with weight of min $[m(r), m(r + \boldsymbol{b}_1)]$ for each upper mask pattern in Figure 5. However, the above GLAC feature dimension (concering $\boldsymbol{F_0}$ and $\boldsymbol{F_1}$) is configured by $d = D + 4D^2$. Therefore, the calculated d-dimensional action representation vector for the $MHI_{XOY}$ is referred to as $\boldsymbol{GMHI_{XOY}}$. Since the vector $\boldsymbol{g}$ is sparse, the feature vector computation is flexible. One can take a look at the work in [14] for deeper knowledge in GLAC.

## B. Action Representation

The vectors $\boldsymbol{GMHI_{XOY}}$, $\boldsymbol{GMHI_{YOZ}}$ and $\boldsymbol{GMHI_{XOZ}}$ are gained by passing the set of MHI to the GLAC individually. Those three action vectors are fused to a single vector as $\boldsymbol{GMHI} = [\boldsymbol{GMHI_{XOY}}; \boldsymbol{GMHI_{YOZ}}; \boldsymbol{GMHI_{XOZ}}]$ to represent an action with motion image based texture features. In the same way, the vector $\boldsymbol{GSHI} = [\boldsymbol{GSHI_{XOY}}; \boldsymbol{GSHI_{YOZ}}; \boldsymbol{GSHI_{XOZ}}]$ is obtained with the end by end concatenation of the SHIs to describe the action by static image based texture features. Clearly, the $\boldsymbol{GMHI}$ and the $\boldsymbol{GSHI}$ vectors are complementary to each other and thus we combine them to a single vector to represent an action at optimal level. We think that our action representation enhances the discriminating capacity of the proposed system.

## C. Action Classification

It has been turn out that the *l2-regularized Collaborative Representation Classifier (l2-CRC)* is able to exhibit the promising outcome in action labeling by [25], [28]. Hence, the fused vector of $\boldsymbol{GMHI}$ and $\boldsymbol{GSHI}$ is fed into the *l2-CR C* to predict the class of a new action sample. However, for an action dataset with $C$ action categories, the $l_2$-CRC consists of a dictionary of $\boldsymbol{n}$ training vectors as $\boldsymbol{P} = [\boldsymbol{P_1}, \boldsymbol{P_2}, \dots \dots, \boldsymbol{P_C}] = [\boldsymbol{p_1}, \boldsymbol{p_2}, \dots \dots, \boldsymbol{p_n}] \epsilon R^{D \times N}$, here $D$ is the length of a sample and $N$ stands for the total number of training action vectors. Also, $P_j \in R^{D \times M_j}, (j = 1,2, \ldots \ldots, C)$ is used to indicate a set of training action vectors with the class label $j$ and $p_i \epsilon R^D (i = 1,2, \ldots, n)$ denotes the $i^{th}$ action vector. However, a test action vector $S \epsilon R^D$ could be figured out as

$$S = P\beta \tag{7}$$

In the above equation, $\beta$ is the $N$-dimensional column vector corresponding to the coefficients which are equal to the training vectors. As described in [55], solution of Equation (7) is not possible directly and hence the equation is solved through the following optimization problem.

$$\hat{\beta} = \frac{\arg min}{\beta} \{\|S - P\beta\|_2^2 + \mu \|A\beta\|_2^2\} \tag{8}$$

In Equation (8), $A$ is known as Tikhonov regularization matrix [18] and $\mu$ is the regularization parameter. The tuning of $\mu$ is very important to get the optimal action classification. The term involved with $A$ ensures the employment of the former information of the solution by utilizing the technique as discussed in [56], [57], [58]. The training vectors that are not close to the test vector are allocated less weight than the vectors that are very similar to the test sample. Finally, the matrix $A \in R^{D \times N}$ is configured as

$$A = \begin{bmatrix} \|v - p_1\|_2 & 0 & \cdots & 0 \\ 0 & \|v - p_2\|_2 & & 0 \\ \vdots & & \ddots & \vdots \\ 0 & & \cdots & \|v - p_n\|_2 \end{bmatrix} \tag{9}$$

By [59], $\hat{\beta}$ could be determined as follows:

$$\hat{\beta} = (P^T P + \mu A^T A)^{-1} P^T S \tag{10}$$

Now, $\hat{\beta}$ can be decomposed into $C$ subsets with the category of the training vectors and which can be expressed as $\hat{\beta} = [\widehat{\beta_1}; \widehat{\beta_2}; \widehat{\beta_3}; \ldots \ldots \ldots; \widehat{\beta_C}]$. Eventually, the class label of the test vector $S$ is evaluated by

$$class\ (S) = \frac{\arg min}{j \in \{1,2, \ldots \ldots, C\}} \{q_j\}, \tag{11}$$

where $q_j$ is defined by:

$$q_j = \left\|S - P_j \widehat{\beta_j}\right\|_2. \tag{12}$$

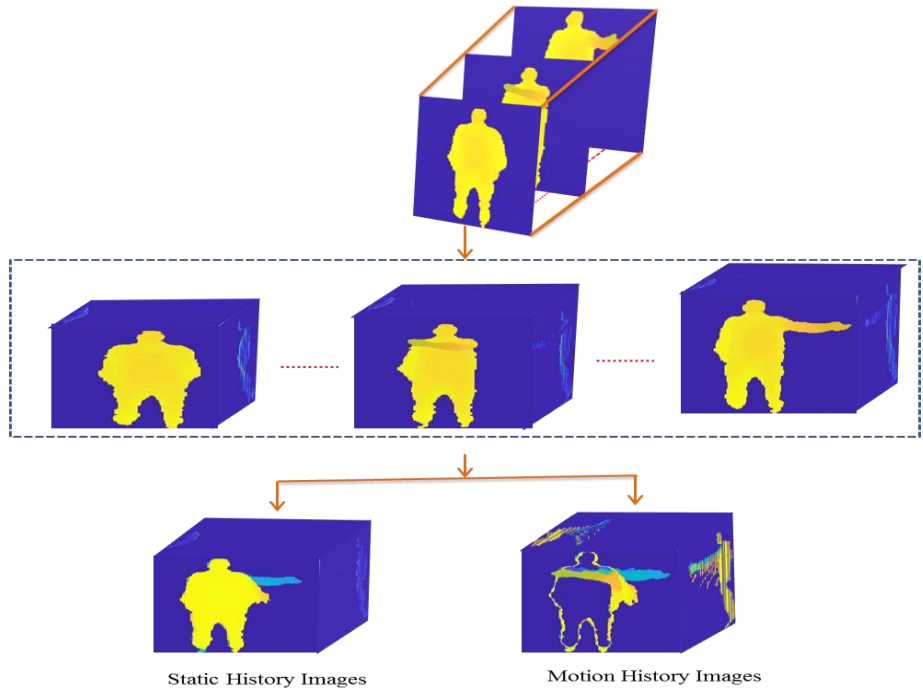

Figure 4. Example of SHIs and MHIs generation using the 3DMTM corresponding to the *horizontal wave* action

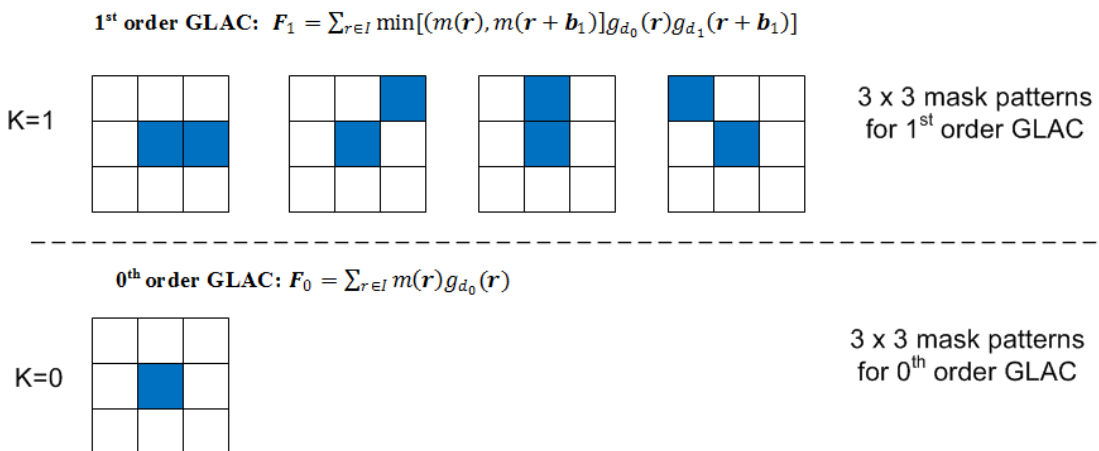

Figure 5. GLAC computing mask patterns for $K\epsilon\{0,1\}$

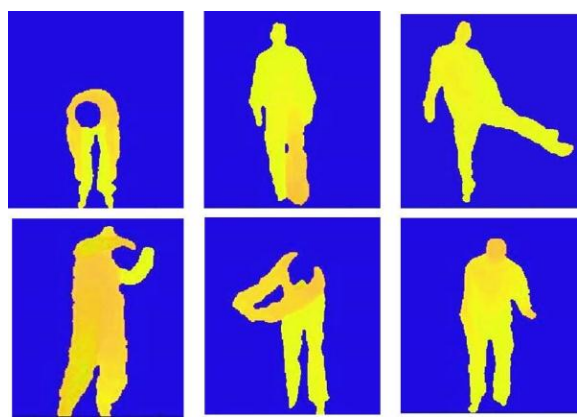

Figure 6. Example of action snaps in *MSR-Action3D* dataset

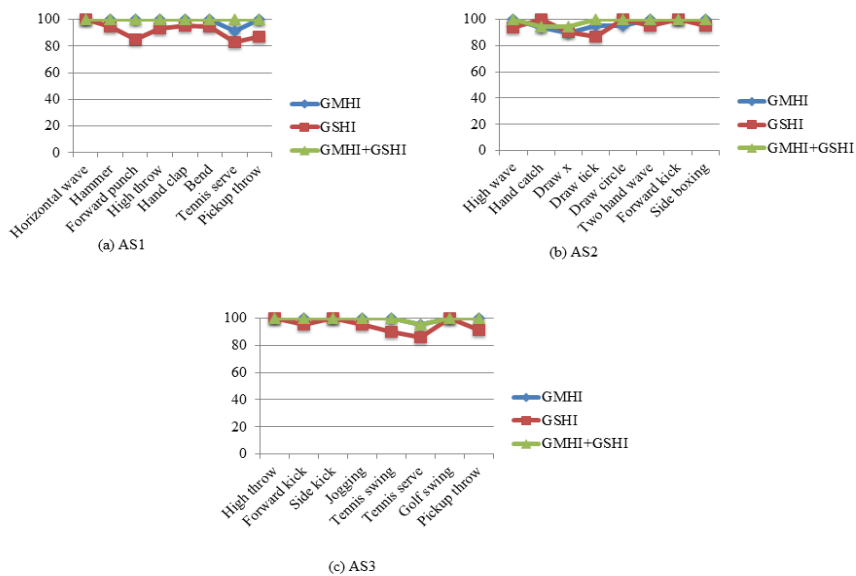

Figure 7. Class specific accuracies in *test one*

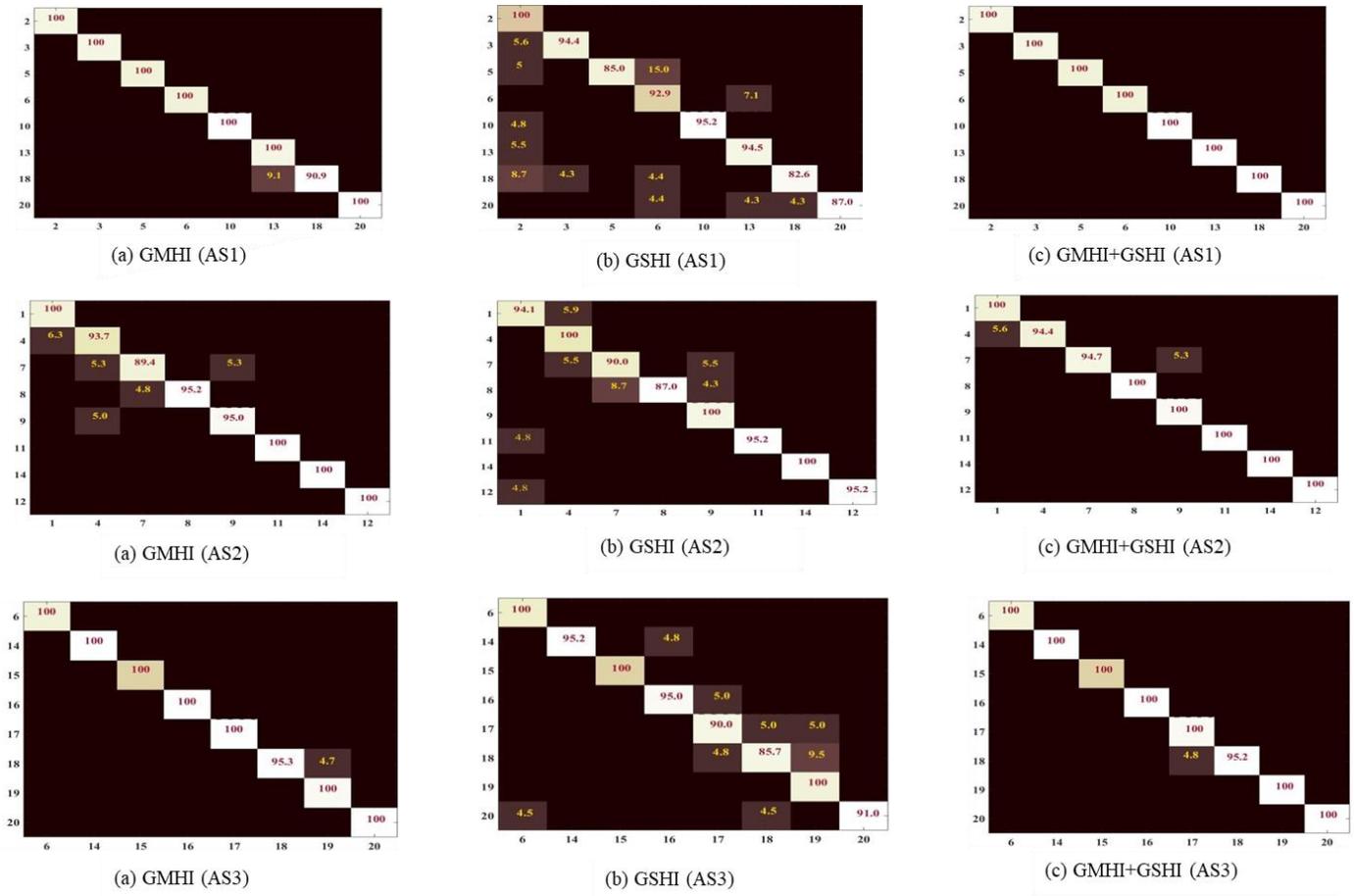

Figure 8. Confusion matrices in *test one*

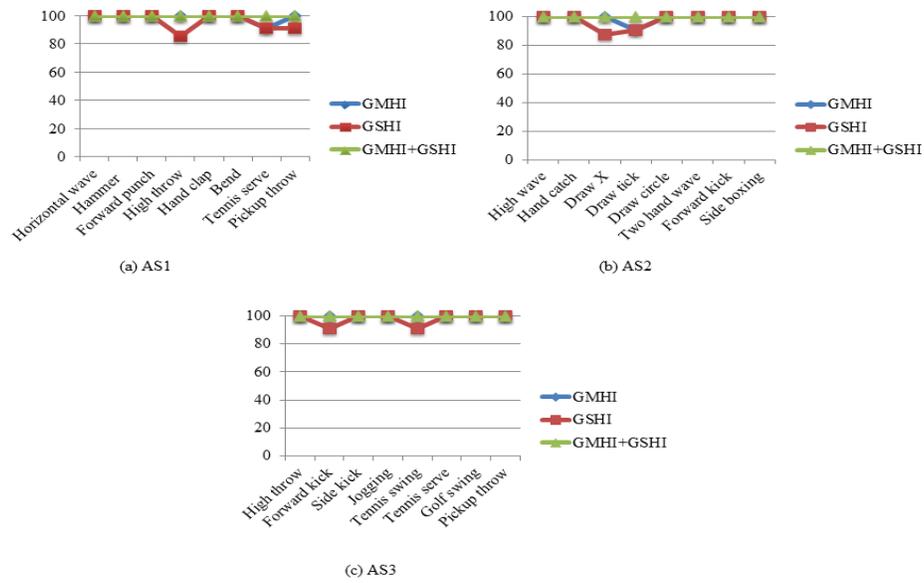

Figure 9. Class specific accuracies in *test two*

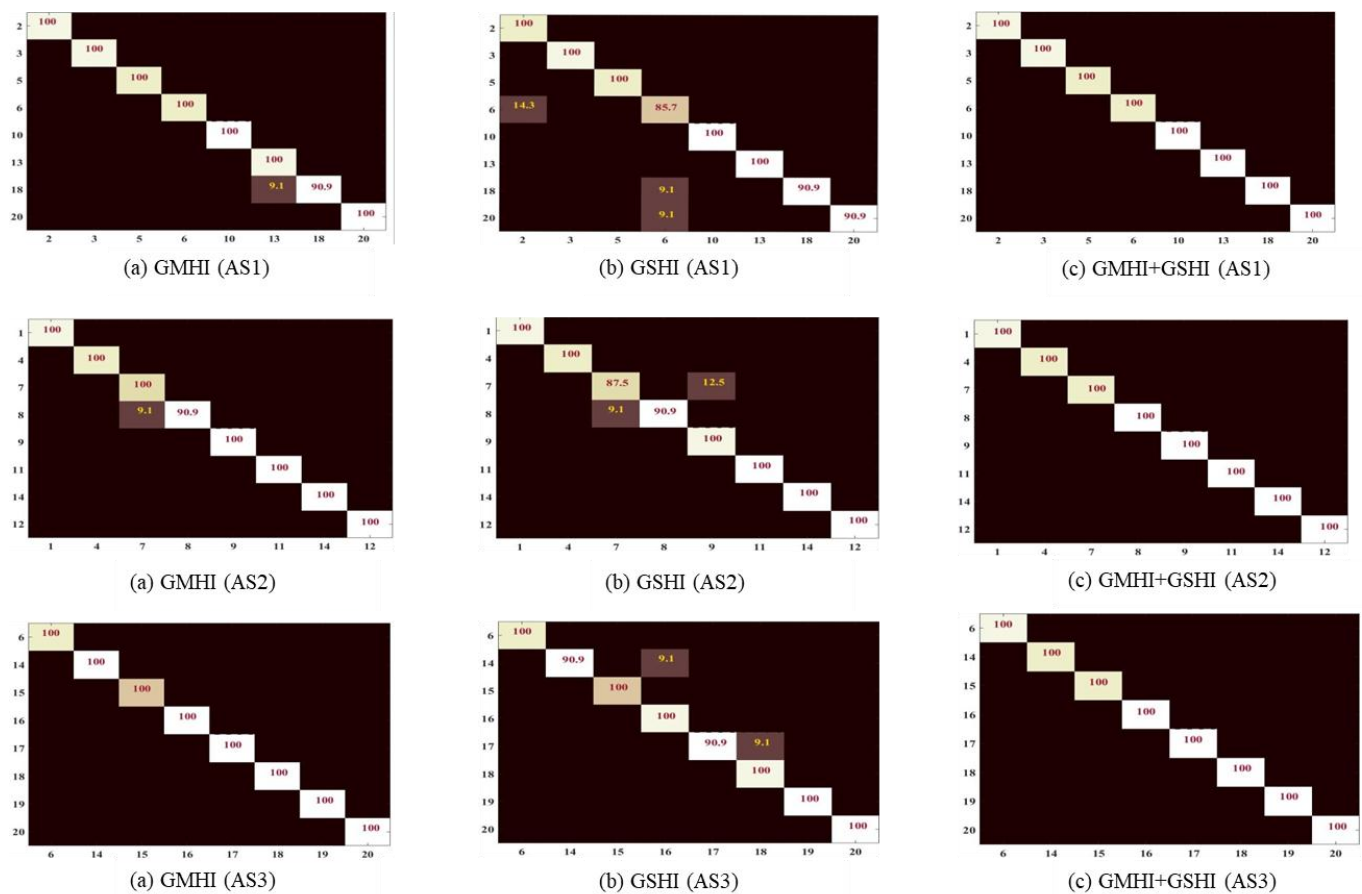

Figure 10. Confusion matrices in *test two*

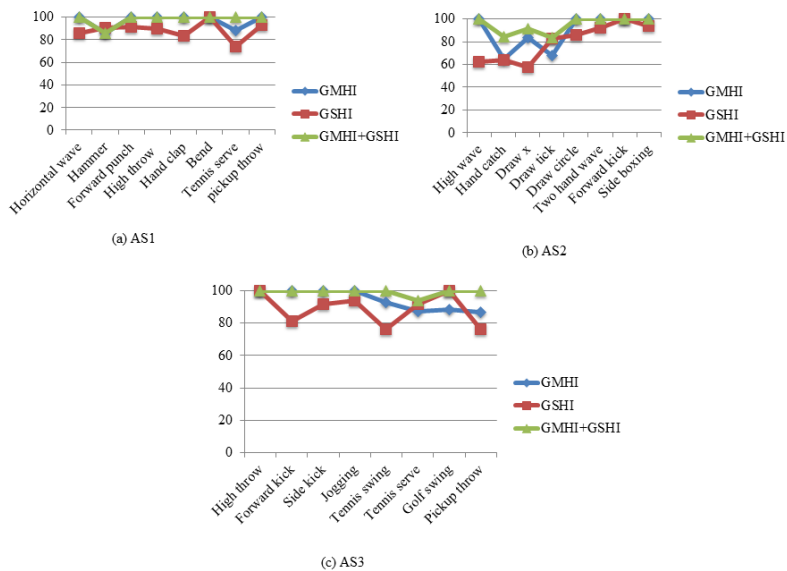

Figure 11. Class specific accuracies in *cross subject test*

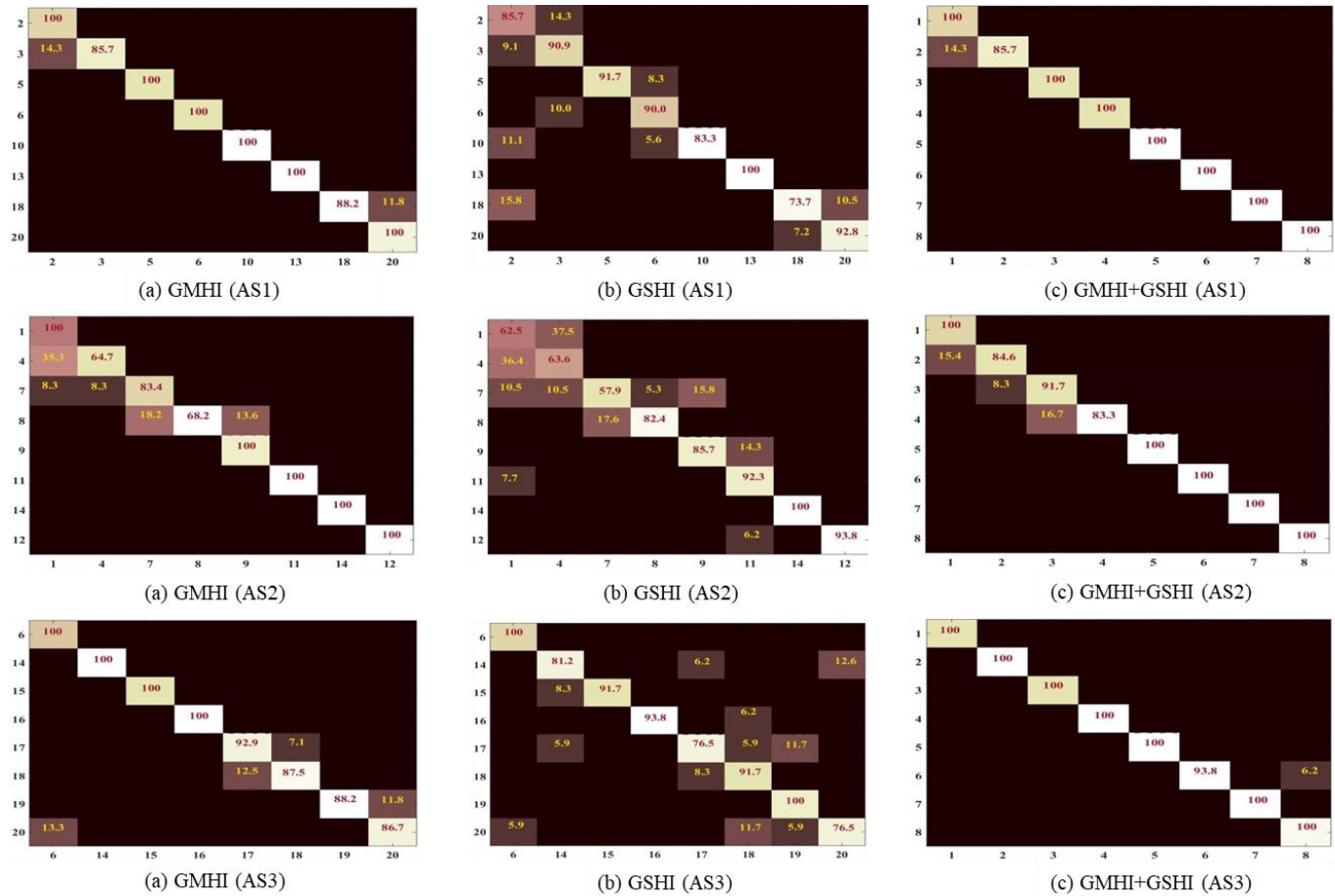

Figure 12. Confusion matrices in *cross subject test*

## IV. EXPERIMENT

The introduced approach is tested on a CPU platform with an Intel i7-4790 Quad-core CPU @3.60 GHz and a RAM of 8GB. Our system is evaluated on the MSRAction3D [15], DHA [16], and UTD-MHAD [45] action samples comprehensively.

### A. Results on the MSRAction3D Dataset

**MSRAction3D dataset** [15] is organized with samples of 20 action categories. Each sample video was taken by 2 or 3 times from 10 individuals with a depth camera. The contents of the dataset are of **: "high wave (1)", "horizontal wave (2)", "hammer (3)", "hand catch (4)", "forward punch (5)", "high throw (6)", "draw x (7)", "draw tick (8)", "draw circle (9)", "hand clap (10)", "two hand wave (11)", "side boxing (12)", "bend (13)" "forward kick (14)", "side kick (15)", "jogging (16)", "tennis swing (17)", "tennis serve (18)", "golf swing (19)", "pick up and throw (20)"**. The dataset is actually contains the gaming action videos with inter-class similarity issue as observed between *draw x* and *draw tick*. We adopt the similar assessment setup by following [12],[15],[18]and[22] to have a fair comparison of the proposed system. In Figure 6, a couple of depth maps are shown as action example of the dataset.

### First Experimental Setup and Results:

In this setup, the entire 20 action classes are put into three sets (see Table I). We carry out *test one, test two and cross subject test* on each action set according to the work in [15],[18]. In GLAC, we always employ the Roberts operator to calculate the gradient vector in MHI /SHI as the Roberts operator is the most compact and most effective compared to other operators [14]. In the whole evaluation, the GLAC parameters' pair $(D, \Delta r)$ is set to (8,1) to operate the descriptor on the MHIs and SHIs with 5-fold cross validation strategy. The spatial bin parameter is tuned as $b_s = 1 \times 2$ similarly. The *l2-CRC* parameter $\mu$ is also tuned to 0.0001 in the same manner. In this setting, the dimension of an action representation vector is 3168 and thus, our work employs the Principle Component Analysis (PCA) to improve the classification performance of the system. In PCA, the principle components which are account for 99% of the entire variation are retained. The dimension of feature vectors, after using the PCA algorithm, is shown in the corresponding accuracy table of each test case.

TABLE I: THREE SUBSETS OF THE MSR-ACTION 3D DATASET

| Label | AS1 | Label | AS2 | Label | AS3 |
|-------|-----|-------|-----|-------|-----|
| 2 | Horizontal wave | 1 | High wave | 6 | High throw |
| 3 | Hammer | 4 | Hand catch | 14 | Forward kick |
| 5 | Forward punch | 7 | Draw x | 15 | Side kick |
| 6 | High throw | 8 | Draw tick | 16 | Jogging |
| 10 | Hand clap | 9 | Draw circle | 17 | Tennis swing |
| 13 | Bend | 11 | Two hand wave | 18 | Tennis serve |
| 18 | Tennis serve | 14 | Forward kick | 19 | Golf swing |
| 20 | Pickup throw | 12 | Side boxing | 20 | Pickup throw |

**In *test one***, 1/3 action samples relevant to all the action classes in each action set are employed in training set and the rest sample of those classes of the set are taken in test set. The alone and average classification outcomes concerning the AS1, AS2 and AS3 are depicted in Table II. The table indicates the introduced approach (i.e., the combination of ***GMHI*** and ***HI***) attains 99.34% average recognition accuracy. The confusion matrices in this test are represented in Figure 8. The confusion matrix contains information about actual and predicted labels as done by the classifier and thus, summarizes the performance of a classification algorithm. In our case, rows of each confusion matrix represent the actual labels for actions and columns represent the labels predicted by the classifier.

However, our recognition method cannot reach classification accuracy of 100% as the *hand catch*, *draw x* and *tennis serve* actions are confused with *high wave, draw circle and tennis swing* respectively (see Figure 8). However, the confusion/misclassification rates are very small and confusion mostly happens for recognizing analogous actions.

The classification outcome by ***GMHI*** alone and by ***GSHI*** alone are also reported in the table. Their recognition accuracies are not promising as the proposed method. Only the ***GMHI*** based system gains an equivalent result (which is 99.33%) to the proposed method for the AS3. It is clear that the recognition accuracies using the ***GMHI*** features outperform the accuracies based on the ***GSHI*** features. Although the ***GMHI*** achieves superiority over the ***GSHI*** features, but the ***GMHI*** features cannot recognize all actions in the dataset. Besides, the ***GSHI*** features are not showing inferior results for all actions compared to the ***GMHI*** features. Overall, the ***GMHI*** features may achieve the promising recognition performance for several specific actions and the ***GSHI*** features can be inferior (for certain action, the reverse may also happen). For example, in Figure 7, the ***GMHI*** features show the higher recognition accuracy than the ***GSHI*** features to classify the *high wave* action in AS2. On the other hand, the *hand catch* action in AS2 is recognized more accurately using the ***GSHI*** features than using ***GMHI*** features. From above, it is clear that these two descriptors are complementary enough and thus, their fusion may improve the recognition accuracy considerably. As a result, by way of merging the ***GMHI*** vectors and the ***GSHI*** vectors, the whole recognition outcome is upgraded significantly on the situations while utilizing the ***GMHI*** features only or the ***GSHI*** features only. As instance, the proposed approach has 1% more recognition rate ***GMHI*** than and 9% more recognition rate than ***GSHI*** on AS1. It noticeably demonstrates the benefit of fusing these features for increasing the recognition accuracy. To further investigate the enhancement, we observe comparison regarding individual action type recognition accuracy relevant to the introduced approach, ***GMHI*** and ***GSHI*** in all the action sets in Figure 7. As obvious in the figure, the introduced fusion approach is capable to increase the action categorization accuracy for a greater portion of the twenty actions, e.g., *hammer, draw tick,* and *draw x*. It should be noted that the same action could be confused in the proposed method as well as in ***GMHI*** and ***GSHI***. But the confusion rate is comparatively lower in the fusion method.

TABLE II: RECOGNITION RESULTS ON THE THREE ACTION SETS IN TEST ONE

| | | GMHI | GSHI | GMHI + GSHI | Feature dimension |
|---|---|------|------|-------------|-------------------|
| **Test One** | AS1 | 98.7 | 90.7 | **100** | 70 |
| | AS2 | 96.7 | 94.7 | **98.69** | 37 |
| | AS3 | **99.33** | 94.00 | **99.33** | 66 |
| | Average | 98.24 | 93.13 | **99.34** | - |

***In Test Two***, the *l2-CRC* employs 2/3 action samples as training samples of the entire samples in every action set. The remaining samples of the set are engaged in the test stage. The classification outcomes for the ***GMHI***, ***GSHI*** and their combination version are illustrated by Table III. In the table, it can be seen for all the action sets, our proposed system exhibits superiority over the ***GMHI*** and ***GSHI*** methods notably. The introduced approach gains 100% classification rate for the three sets whereas the ***GMHI*** and ***GSHI*** cannot do it. The ***GMHI*** based approach shows 100% recognition outcome for AS3 only. The recognition accuracy for every class can be found in Figure 9 to become clearer about the supremacy of the proposed approach over the ***GMHI*** and the ***GSHI*** approaches. Besides, confusion between two actions are absent in the proposed method where confusions are occurred in ***GMHI*** and ***GSHI*** methods (see Figure 10).

TABLE III: RECOGNITION RESULTS ON THE THREE ACTION SETS IN TEST TWO

|  |  | GMHI | GSHI | GMHI + GSHI | Feature dimension |
|---|---|---|---|---|---|
| **Test Two** | AS1 | 98.6 | 95.9 | **100** | 151 |
|  | AS2 | 98.7 | 97.3 | **100** | 137 |
|  | AS3 | **100** | 97.3 | **100** | 149 |
|  | Average | 99.1 | 96.83 | **100** |  |

***In Test Three/Cross Subject Test***, the corresponding action samples of the individuals 1, 3, 5, 7, and 9 are used in the *l2-CRC* training samples and the samples from actors 2, 4, 6, 8, and 10 are passed to the *l2-CRC* as query samples. Form Table IV, it can be seen that the introduced method attains enhanced results for all aspects than others. More precisely, the proposed system outperforms ***GMHI*** by 7 % and ***GSHI*** by 14%. It should be noted that our method couldn't gain 100% recognition accuracy by avoiding the confusion between similar actions (see Figure 12) but the fusion method leads lower confusion compared to ***GMHI*** and ***GSHI*** methods. Figure 11 demonstrates the class-wise recognition outcomes for all the approaches.

TABLE IV: RECOGNITION RESULTS ON THE THREE ACTION SETS IN *CROSS SUBJECT TEST*

|  |  | GMHI | GSHI | GMHI + GSHI | Feature dimension |
|---|---|---|---|---|---|
| **Cross Subject Test** | AS1 | 96.23 | 87.74 | **98.11** | 40 |
|  | AS2 | 86.73 | 80.53 | **94.69** | 33 |
|  | AS3 | 93.75 | 87.5 | **99.11** | 65 |
|  | Average | 92.23 | 85.25 | **97.3** |  |

The recognition outcomes of the proposed system are also compared with methods which were assessed on MSR-Action3D dataset through following the same experimental mannerisms. Table V exhibits the comparison in average recognition accuracy (%) for all the test strategies. Note that, the table includes those methods which were validated on MSR-Action3D dataset by analogous experimentations. The maximum classification outcome is focused by bold face. It is can be observed, our system attains supremacy over all the systems listed in the table. Furthermore, the proposed approach exhibits the state-of-the-art recognition rates of **99.34 %, 100%** and **97.3%** in the *test one*, *test two* and *cross subject test* respectively. Especially for the most challenging *cross subject test*, the proposed approach beats the listed methods significantly, leading to 4.1 % improvement over the second highest accuracy (93.2 % in [42]). In addition, the recognition system shows superiority over the deep learning systems reported in [50].The action classification results based on ***GMHI*** and ***GSHI*** methods are also represented in the table.

TABLE V: AVERAGE ACCURACY COMPARISON ON THE MSR ACTION 3D DATASET ON THE FIRST SETTING

| Methods | Test One (%) | Test Two (%) | Cross Subject Test (%) |
|---|---|---|---|
| Bag of 3D Points[15] | 91.6 | 94.2 | 74.7 |
| DMM-HOG [18] | 95.8 | 97.4 | 91.6 |
| DMM [25] | 97.4 | 99.1 | 90.5 |
| DMM-LBP-FF[26] | 98.7 | **100** | 94.9 |
| DMM-LBP-DF[26] | 98.2 | **100** | 94.7 |
| STOP[22] | 96.8 | 98.3 | 87.5 |
| HOJ3D[14] | 96.2 | 97.2 | 79.0 |
| Skeletons Lie Group [47] | - | - | 92.5 |
| Evolutionary Joint Selection [42] | - | - | 93.2 |
| MS[50] | 93.6 | 94.3 | 86.3 |
| SMF[50] | 96.7 | 98.7 | 89.1 |
| BDL[50] | 94.1 | 95.6 | 87.6 |
| SMF-BDL[50] | 97.3 | 99.1 | 90.8 |
| **Our Method (GMHI + GSHI)** | **99.34** | **100** | **97.3** |

### Second Experimental Setup and Results

The evaluation method followed by [12], [22] is also employed here to have a fair comparison. In this setup, we employ all the action classes instead of splitting them into several sets of action classes. The action samples taken by the persons of index 1, 3, 5, 7, 9 are utilized for passing to the classifier as training samples and the samples about the rest subjects are used as test samples. We use $D = 8$, $\Delta r = 1$, $b_s = 1 \times 3$ and $\mu = 0.0001$ as optimal values by setting them with the similar technique as discussed in the first experimental setup. The length of feature vector is shrinking to 4752 to 85 by the PCA. Table VI presents the recognition accuracy based comparison of our method with other methods with same evaluation strategy. The table contains recognition systems which were tested on the same dataset and same experimental setup. We also compare our system with the deep structured learning system described in [51]. Our method outperforms the method 2D-CNN [51] by 3.3% and the method 3D-CNN [51] by 8.4%. The comparative classification outcomes can be found in their relevant papers. From Table VI, It should be noted that our system significantly exhibits supremacy over the deep learning based methods. Figure 13 illustrates the confusion matrix of this setup.

TABLE VI: RECOGNITION ACCURACY COMPARISON ON THE MSR ACTION 3D DATASET ON THE SECOND SETTING

| Method | Accuracy (%) |
|---|---|
| Random Occupancy Pattern [23] | 86.5 |
| DMM-HOG [18] | 88.7 |
| HON4D [35] | 88.9 |
| DSTIP [19] | 89.3 |
| Moving Pose [43] | 91.7 |
| Actionlet Ensemble [46] | 88.2 |
| Skeletons Lie group [47] | 89.5 |
| Skeletal Quads [44] | 89.9 |
| Super Normal Vector [12] | 93.1 |
| 2D-CNN[51] | 91.2 |
| 3D-CNN[51] | 86.1 |
| HOG3D+LLC [37] | 90.9 |
| Hierarchical 3D Kernel [38] | 92.7 |
| DMM-LBP-DF [26] | 91.9 |
| DMM-LBP-DF [26] | 93.0 |
| HP-DMM-CNN [66] | 92.3 |
| BoA [66] | 86.9 |
| **Our Method (GMHI + GSHI)** | **94.5** |

### B. Results on the DHA dataset

**DHA dataset** [16] includes some actions of Weizmann dataset [60]. The DHA dataset has samples of 23 action classes where the descriptions of samples of 1 to10 categories are similar to the Weizmann dataset [61]. The 23 types are: **"arm-curl (1)", "arm-swing (2)", "bend (3)", "front-box (4)", "front-clap (5)", "golf-swing (6)", "jack (7)", "jump (8)", "kick (9)", "leg-cur (10)", "leg-kick (11)", "one-hand-wave (12)", "pitch (13)", "pjump (14)", "rod-swing (15)", "run (16)", "skip (17)", "side (18)", "side-box (19)", "side-clap (20)", "tai-chi (21)", "two-hand-wave (22)", "walk (23)"**. There are 483 action samples in the dataset. Those samples are recorded from 21 persons (12 males and 9 females). The dataset is challenging enough since different action classes have similar motions such as *leg-curl* and *leg-kick*, *run* and *walk*, etc. Example of sample depth action frames of multiple human actions in the action dataset is illustrated in Figure 14. The set of action samples captured from performers 1, 3, 5, 7, 9, 11, 13, 15, 17, 19, and 21 are involved in the training session and the samples obtained from the remaining actors are engaged in test state. For this dataset, $D = 8$, $\Delta r = 1$, $b_s = 1 \times 2$ and $\mu = 0.0001$ are chosen by the earlier discussed method. The dimension of feature vector is shortened from 3168 to 252 by the PCA algorithm. For DHA dataset, the comparison of accuracy for different techniques is shown in Table VII, with the techniques evaluated on the same dataset. In Table VII, it can be seen that our system attains a remarkable recognition rate of 99.1%. From Figure 15, we can observe, the proposed method classifies 21 actions among 23 by the accuracy of 100%. Moreover, the comparison of our system with other systems by similar experimental setup demonstrates that our system achieves outstanding outcomes over all the methods included in Table VII.

TABLE VII: RECOGNITION ACCURACY COMPARISON ON THE DHA DATASET

| Method | Accuracy (%) |
|--------|--------------|
| D-STV/AS[16] | 86.8 |
| SDM-BSM [36] | 89.5 |
| D-DMHI-PHOG [45] | 92.4 |
| DMPP-PHOG [45] | 95.0 |
| DMM-LBP-DF [26] | 91.3 |
| DMMs-FV [33] | 95.4 |
| 3DHoT-MBC [34] | 96.7 |
| **Our Method** **(GMHI + GSHI)** | **99.1** |

### C. Results on UTD-MHAD Dataset

The **UTD-MHAD** [17] dataset includes 861 action samples of 27 action classes. All action samples are generated by 8 persons (4 females and 4 males) where everybody takes 4 trials for each action class. The list of 27 classes are : **"right arm swiping to the left (1)"**, **"right arm swiping to the right (2)"**, **"right hand wave (3)"**, **"two hand front clap (4)"**, **" right arm throw (5)"**, **"cross arms in the chest (6)"**, **"basketball shoot (7)"**, **"right hand draw x (8)"**, **"right hand draw circle (clockwise) (9)"**, **"right hand draw circle (counter clockwise) (10)"**, **"draw triangle (11)"**, **"bowling (right hand) (12)"**, **"front boxing (13)"**, **"baseball swing from right (14)"**, **"tennis right hand forehand swing (15)"**, **"arm curl (two arms) (16)"**, **"tennis serve (17)"**, **"two hand push (18)"**, **"right hand knock on door (19)"**, **"right hand catch an object (20)"**, **"right hand pick up and throw (21)"**, **"jogging in place (22)"**, **"walking in place (23)"**, **"sit to stand (24)"**, **"stand to sit (25)"**, **"forward lunge (left foot forward) (26)"**, **"squat (two arms stretch out) (27)"**. The dataset considers diverse action classes such as sport actions (e.g., *bowling*), hand gestures (e.g., *draw x*), daily activities (e.g., *knock on door),* and training exercises (e.g., *arm curl).* Some example depth maps of the dataset are figured out in Figure 16. The samples provided by the players 1, 3, 5, and 7 are included in the training set and the samples captured from residual actors are placed in the test set. The system uses $D = 8$, $\Delta r = 1$, $b_s = 3 \times 5$ and $\mu = 0.0001$ to obtain the expected outcome. The length of feature vector is reduced from 23760 to 94 by PCA. The comparison between our system and other existing systems (evaluated on UTD-MHAD dataset) are shown in Table VIII. For the table, it can be seen, the proposed method attains higher recognition accuracy of 5.1% than the best existing approach (the accuracy of the indicated algorithm is 84.4%) stated in [34]. The method as described in [67] outperforms our method by 2.2% but it is noticeable that the method in [67] was evaluated on the RGB action data whereas our method uses depth data. In fact, this type of comparison is not fair although we consider it in our work. For more clarification of the performance of our method, the confusion matrix is illustrated in Figure 17.

TABLE VIII: RECOGNITION ACCURACY COMPARISON ON THE UTD-MHAD DATASET

| Method | Accuracy (%) |
|--------|--------------|
| Adaboost.M2 [52] | 83.0 |
| DMM-HOG [18] | 81.5 |
| Kinect[17] | 66.1 |
| Inertial[17] | 67.2 |
| Kinect & Inertial [17] | 79.1 |
| 3DHoT-MBC [34] | 84.4 |
| HP-DMM-CNN [66] | 82.8 |
| BoA [66] | 85.4 |
| Optical flow CNN [66] | 82.6 |
| Structural Average Curves [67] | **91.7** |
| **Our Method** **(GMHI + GSHI)** | 89.5 |

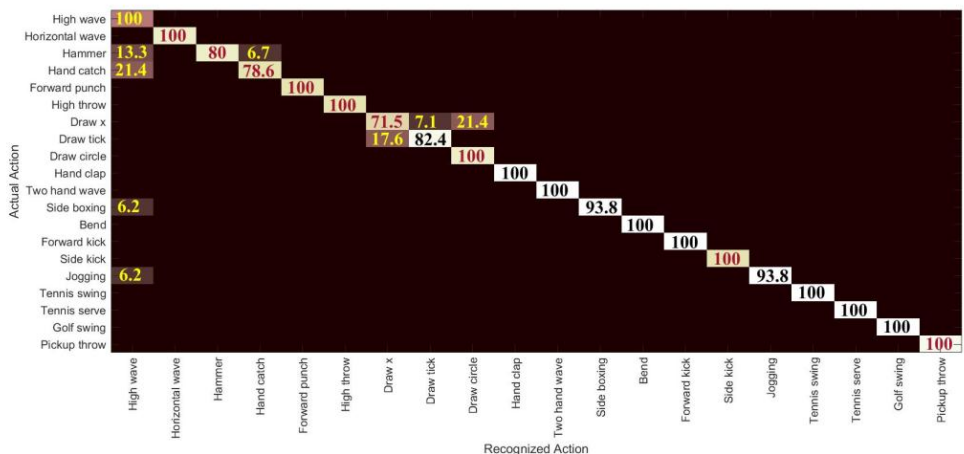

Figure 13. Confusion matrix on the MSRAction3D dataset on setting 2

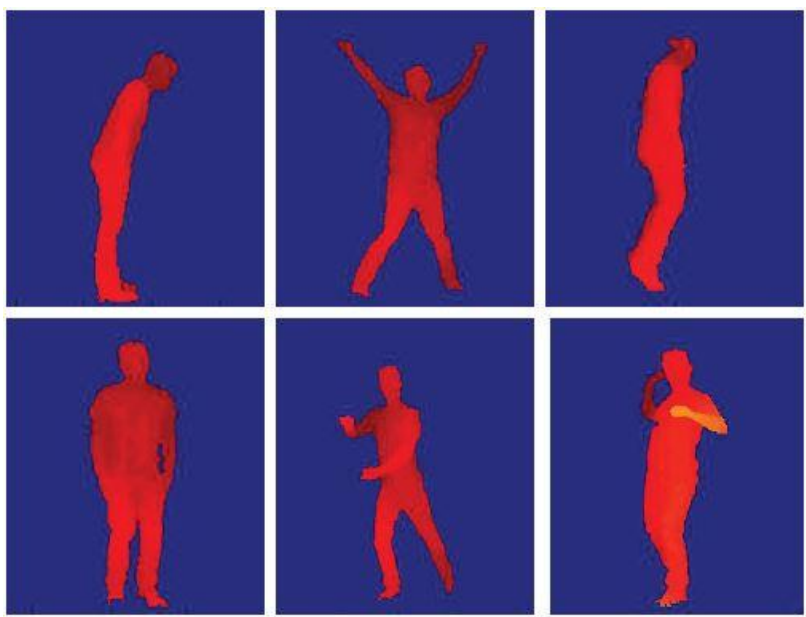

Figure 14. Sample depth images of different actions from the action dataset DHA

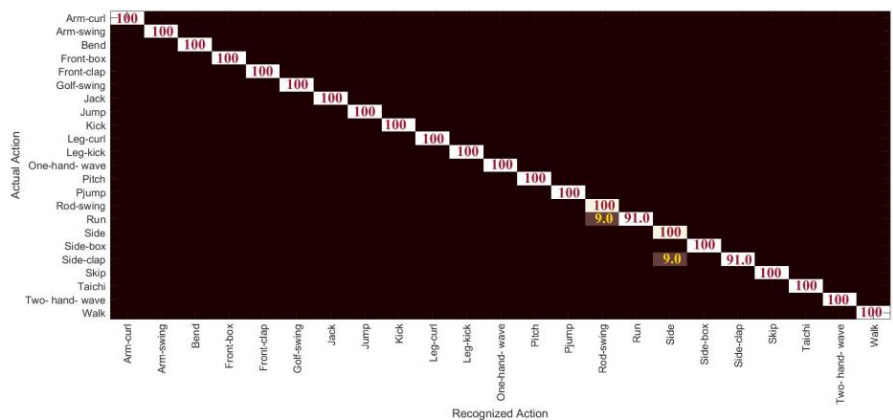

Figure 15. Confusion matrix on the DHA dataset

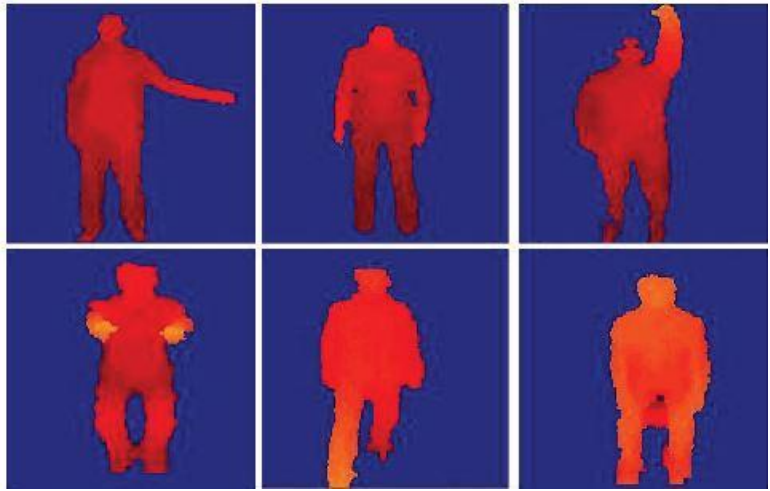

Figure 16. Sample depth images of different actions from the dataset UTD-MHAD

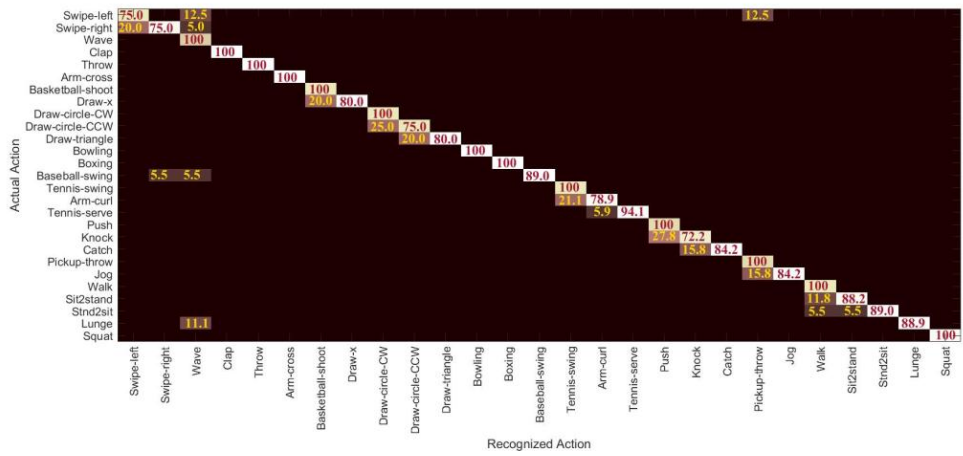

Figure 17: Confusion matrix on the UTD-MHAD dataset

V. CONCLUSION

This paper has mainly proposed an effective action representation strategy by jointly using two sets of features. The system fuses the motion image based texture features (*i.e.*, *GMHI*) with the static image based texture features (*i.e.*, *GSHI*) to represent a depth action with optimal discriminary power. The *GMHI* is computed by passing the MHIs to the GLAC algorithm and the *GSHI* is gained by operating the GLAC on the SHIs. Experimental evaluations are carried out based on three popular action datasets such as *MSR-Action3D*, *DHA*, and *UTD-MHAD*. The evaluation on the *MSR-Action3D* dataset is considered with two different experimental strategies. In the first strategy, the proposed method provides the recognition accuracy of **97.3%** for the most challenging *cross subject test*. In the second strategy, the considerable recognition outcome is of **94.5%.** In both settings, the recognition results are compared with the results based on hand-crafted features as well as deep learning models. However, the proposed method shows superiority over them. Moreover, the introduced system shows **99.1%** recognition rate for the *DHA* dataset and **89.5%** recognition rate for the *UTD-MHAD* dataset. Those outcomes could be considered as remarkable since both are very complex datasets. Overall experimental results for the datasets revealed that the proposed system consistently outperforms the other reported methods by achieving the state-of-the-art accuarcy.

The confusion matrix of each experimental result indicates that the proposed method may still face challenge to reduce the confusion between two similar motion pattern actions such as *Draw x* and *Draw tick; Draw x* and *Draw circle* (see Figure 13). The computed MHI/SHI contains somewhat similar motion patterns for two different action classes due to apparent similarity of some depth action images for those actions. For example, some portions of the MHI/SHI for *Draw x* and *Draw tick* actions are similar and thus the confusion is observed between them in Figure 13. This can make an interesting future task. In the future work, to improve the action representation through the MHI and SHI, we plan to split an action video into multiple video segments and to construct MHI and SHI for each segment. The technique implies a number of MHI and SHI corresponding to an action video and will capture a more appropriate description as compared to the use of a single MHI and SHI. Besides, the future scheme may provide sufficient MHI and SHI images for each action to train a deep learning model. As a result, we also aim to build a 2D deep learning model, e.g., 2D CNN, to feed those 2D images to the model to recognize human action more robustly.